\documentclass[letterpaper]{article} % DO NOT CHANGE THIS
\usepackage{aaai24}  % DO NOT CHANGE THIS
\usepackage{times}  % DO NOT CHANGE THIS
\usepackage{helvet}  % DO NOT CHANGE THIS
\usepackage{courier}  % DO NOT CHANGE THIS
\usepackage[hyphens]{url}  % DO NOT CHANGE THIS
\usepackage{graphicx} % DO NOT CHANGE THIS
\usepackage{natbib}  % DO NOT CHANGE THIS AND DO NOT ADD ANY OPTIONS TO IT
\usepackage{caption} % DO NOT CHANGE THIS AND DO NOT ADD ANY OPTIONS TO IT
\usepackage{algorithm}
\usepackage{multirow}
\usepackage{multicol}
\usepackage{bm}
\usepackage{color}
\usepackage{mathrsfs}
\usepackage{latexsym}
\usepackage{epsfig}
\usepackage{booktabs}
\usepackage{siunitx}
\usepackage{amstext}
\usepackage{array}
\usepackage{enumerate}
\usepackage{amsmath} 
\usepackage{latexsym} 
\usepackage{amssymb} 
\usepackage{amstext}
\usepackage{algpseudocode}

\newcommand{\down}[1]{\tiny\textcolor{red}{#1}}

\makeatletter
\def\thanks#1{\protected@xdef\@thanks{\@thanks
\protect\footnotetext{#1}}}
\makeatother
\usepackage{newfloat}
\usepackage{listings}
\DeclareCaptionStyle{ruled}{labelfont=normalfont,labelsep=colon,strut=off} % DO NOT CHANGE THIS
\floatstyle{ruled}
\newfloat{listing}{tb}{lst}{}
\floatname{listing}{Listing}
\title{Mutual-modality Adversarial Attack with Semantic Perturbation}
\author{\bf Jingwen Ye \quad %\orcidID{0000-0001-8415-3597} 
Ruonan Yu \quad
Songhua Liu \quad
Xinchao Wang $^{\dagger}$ \thanks{ $^{\dagger}$ Corresponding author.}\\
National University of Singapore\\
{\tt\small jingweny@nus.edu.sg, \{ruonan,songhua.liu\}@u.nus.edu, 
xinchao@nus.edu.sg}
}

\usepackage{bibentry}
\begin{document}

\maketitle

\begin{abstract}
Adversarial attacks constitute a notable threat to machine learning systems, given their potential to induce erroneous predictions and classifications. However, within real-world contexts, the essential specifics of the deployed model are frequently treated as a black box, consequently mitigating the vulnerability to such attacks.
Thus, enhancing the transferability of the adversarial samples has become a crucial area of research, which heavily relies on selecting appropriate surrogate models.
To address this challenge, we propose a novel approach that generates adversarial attacks in a mutual-modality optimization scheme. Our approach is accomplished by leveraging the pre-trained CLIP model. Firstly, we conduct a visual attack on the clean image that causes semantic perturbations on the aligned embedding space with the other textual modality. 
Then, we apply the corresponding defense on the textual modality by updating the prompts, which forces the re-matching on the perturbed embedding space. 
Finally, to enhance the attack transferability, we utilize the iterative training strategy on the visual attack and the textual defense, where the two processes optimize from each other.
We evaluate our approach on several benchmark datasets and demonstrate that our mutual-modal attack strategy can effectively produce high-transferable attacks, which are stable regardless of the target networks. Our approach outperforms state-of-the-art attack methods and can be readily deployed as a plug-and-play solution.
\end{abstract}

\section{Introductions}

With the milestone performances of Deep Neural Networks (DNNs) in numerous computer vision tasks, the efficiency~\cite{Xinyin23DeepCache,Gongfan23DepGraph,Songhua22DD,Xingyi22DeRy} and reliability~\cite{ye2023partial,ye2022learning,ye2022safe} of these techniques become equally important when deployed in the real world. 
However, recent researches~\cite{Goodfellow2014ExplainingAH} have found that such DNNs are vulnerable to adversarial examples. 
This is, through only a small norm of perturbation applied on the original input, the maliciously crafted adversarial samples could cause misclassification or unexpected behavior to machine learning models.

As a crucial assessment of the strength and security of DNNs, various attack algorithms~\cite{Hayes2017LearningUA,Liu2019PerceptualSensitiveGF} have been proposed, achieving relatively high fooling rates. However, the effectiveness of these attacks is largely affected by different conditions, with the black-box setting being the most challenging yet realistic scenario.
In the black-box setting, attackers cannot access the model's parameters and structure, leading to the need for improving the transferability of attacks to arbitrary target networks~\cite{cheng2019improving,dong2018boosting}. The corresponding methods include ensemble-model attacks~\cite{liudelving}, momentum-based attacks~\cite{dong2018boosting}, input transformation-based attacks~\cite{xie2019improving}, and model-specific attacks~\cite{wuskip}. Such methods aim to enhance the transferability of attacks by either exploiting the inherent weaknesses of the target model or exploring the common vulnerabilities of a group of models.

\begin{figure}[t]
\centering
\includegraphics[scale = 0.65]{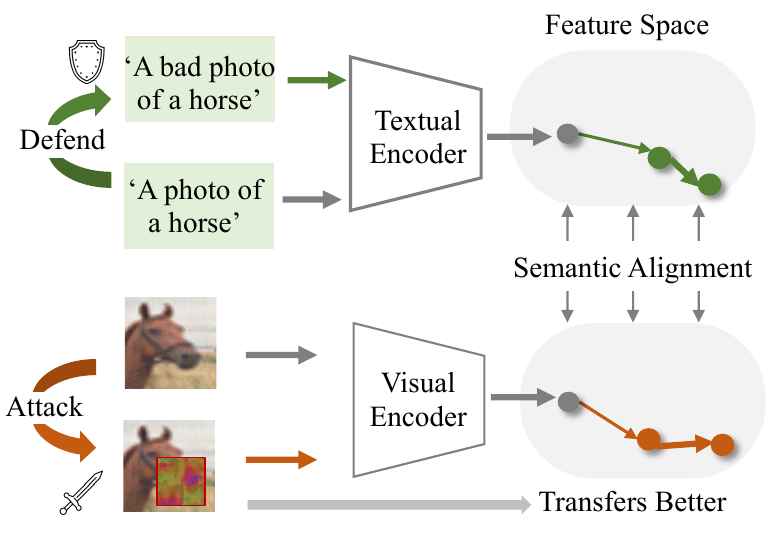}
\vspace{-1em}
\caption{ Joint attack and defense framework in the visual and textual modalities. The visual features are attacked to push away from the original one, while the textual features defend to pull back this similarity gap.
}
\vspace{-1.5em}
\label{fig:intro}
\end{figure}

The majority of current techniques striving to amplify the transferability of adversarial attacks predominantly hinge on the selection of surrogate models. However, these surrogate models often prove to be unstable and profoundly influenced by the architectural similarities between the surrogate model itself and the target networks.
Hence, the careful selection of an optimal surrogate model characterized by a robust feature extractor and exceptional generalizability emerges as a critical factor. Previous studies have used multiple ImageNet-pretrained networks as surrogate models, given the large number of images and object categories in the dataset. 
In this paper, we leverage the recent progress of the CLIP model in computer vision and natural language processing. Having been trained on over 400 million image pairs, CLIP can now serve as our surrogate model, enabling the generation of powerful and broadly effective perturbations.

In addition to its large-scale training data, we utilize the CLIP model as the surrogate model due to its ability to align visual and textual modalities in an aligned feature space. This is accomplished through a visual encoder and textual encoder pairing, allowing us to generate adversarial samples with semantic perturbations. Semantic perturbations differ from previous methods, which simply maximize feature differences. Instead, our approach maximizes semantic differences to ensure that the features after the attack retain explicit semantic information and do not fall into areas without clear semantic meaning, ensuring the effectiveness of the generated attacks.

%In addition, to ensure the attack-generating efficiency, we utilize the generator-oriented approach~\cite{Poursaeed2017GenerativeAP}
%to disrupt low-level features of input images.

In this paper, we propose integrating attack and defense into one framework, building upon the semantic perturbations obtained from the pre-trained CLIP model's aligned visual and textual embedding space. As shown in Fig.~\ref{fig:intro}, we apply visual perturbations to clean images, increasing the semantic difference in the feature space and causing a contradiction with the textual embedding when given the input ``A bad photo of a horse." We then defend against the attack by updating the text prompt template, eliminating this semantic gap and restoring entailment. This iterative attack and defense optimization strategy enhances the attack's transferability to target black-box networks.

To summarize, we make the following contributions:
\begin{itemize}
    \item Firstly, we propose a method to generate reliable adversarial attacks by using the semantic consistency of pre-trained CLIP model to learn perturbations in the semantic feature embedding space. We ensure fidelity by constraining the perturbations with semantic consistency from the text input;
    \item Secondly, we propose an iterative optimization strategy to improve the transferability of the generated attack across different architectures and datasets, where we attack the visual input and defend in the textual one;
    \item Thirdly, we conduct extensive experiments to evaluate the transferability of our proposed approach in cross-dataset, cross-architecture, and cross-task settings. Our results demonstrate that our approach is efficient, effective, and can be used as a plug-and-play solution.
\end{itemize}

%%%%%%%%% BODY TEXT

\section{Related Work}
\subsection{Adversarial Attack}
Adversarial attacks~\cite{Madry2017TowardsDL,Dong2017BoostingAA,guo2019simple,Akhtar2018ThreatOA,Zhang2021ASO} are designed to deceive machine learning models by adding small, imperceptible perturbations to input data, causing the model to generate incorrect outputs or misclassify inputs. One of the traditional attack methods~\cite{Goodfellow2014ExplainingAH} is to use gradient information to update the adversarial example in a single step along the direction of maximum classification loss.
Building on this work, GAMA~\cite{Yuan2021MetaGA} is proposed as a plug-and-play method that can be integrated with any existing gradient-based attack method to improve cross-model transferability.
Besides, many works~\cite{xie2019improving,wang2021enhancing} have been proposed to improve the attack transferability.

To ensure the efficiency of generating attacks, generative model-based attack methods~\cite{Hayes2017LearningUA,Poursaeed2017GenerativeAP,Liu2019PerceptualSensitiveGF,Xiang2022EGMAE,Qiu2019SemanticAdvGA,salzmann2021learning,aichgama} have been extensively studied.
For example, Aishan et al.~\cite{Hayes2017LearningUA} train a generative network capable of generating universal perturbations to fool a target classifier. To generate patch-based attacks, PS-GAN~\cite{Liu2019PerceptualSensitiveGF} is utilized to simultaneously enhance the visual fidelity and attacking ability of the adversarial patch. %As an end-to-end and effective solution for unrestricted adversary example generation, EGM~\cite{Xiang2022EGMAE} contains a conditional reference generator and a conditional adversarial transformer.
%To improve the transferability across datasets and network architectures, GAMA~\cite{aichgama} demonstrates the utility of the CLIP model as an attacker's tool to train formidable perturbation generators for multi-object scenes.

Other than designing attacks in the image domain, attacking NLP models~\cite{morris2020textattack,boucher2022bad,chen2021badnl,zhang2020adversarial,Perez2022IgnorePP} has become a popular research direction. Specifically, prompt learning attacks have attracted many researchers as a lighter method to tune large-scale language models, which can be easily attacked by illegally constructed prompts. Shi et al.~\cite{shi2022promptattack} propose a malicious prompt template construction method to probe the security performance of PLMs. Du et al.~\cite{Du2022PPTBA} propose obtaining poisoned prompts for PLMs and corresponding downstream tasks by prompt tuning. %Perez et al.~\cite{Perez2022IgnorePP} investigate goal hijacking and prompt leaking attacks, demonstrating that even those with low aptitude can easily exploit GPT-3’s stochastic nature, creating long-tail risks.

Different from the above attack methods, we propose the plug and play dynamic updating method, which boosts the transferability. 

\begin{figure*}[t]
\centering
\includegraphics[scale = 0.6]{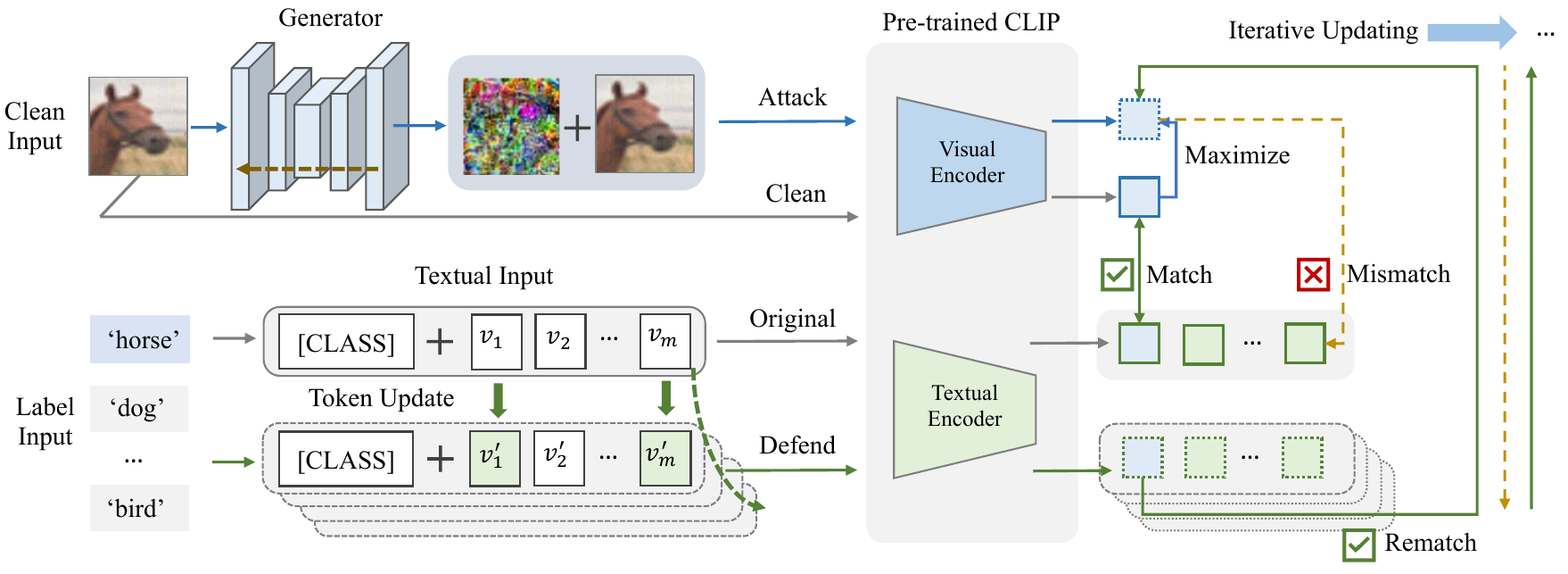}
\caption{ The framework of the joint Mutual-modality attack-defense method. We use the pre-trained CLIP as the surrogate model. The generator is optimized by maximizing the difference with the clean image as input. And the textual input is updated to re-match the features from the textual encoder and the visual encoder.  
}
\label{fig:framework}
\vspace{-1em}
\end{figure*}

\subsection{Vision-and-Language Models}
A vision-and-language model~\cite{jia2021scaling,radford2021learning,mu2022slip,yaofilip,Gongfan23DiffPruning,Xinyin23NeurIPS} is a powerful learning model processing both images and text in a joint manner to align the vision and texture embeddings.
As the most popular VL model, CLIP~\cite{radford2021learning} learn SOTA image representations from scratch on a dataset
of 400 million image-text pairs collected from the internet, which enables various tasks like zero-shot classification. To further boost the VL models' performance, Zhou et al.~\cite{zhou2022learning} propose CoOp to models the prompt’s context words with learnable vectors for adapting CLIP-like vision-language models for downstream image recognition.
And CoCoOp \cite{zhou2022conditional} extends CoOp by further learning a lightweight neural network to generate for each image an input-conditional token.
%In addition, similar visual prompt~\cite{jia2022visual} can be added as an efficient and effective alternative to full fine-tuning for large-scale models in vision.

Target at such large-scale pre-trained VL models, there are a bulk of adversarial attack approaches~\cite{noever2021reading,zhang2022towards} try to fool it.
Noever et al.~\cite{noever2021reading} demonstrate adversarial attacks of spanning basic typographical, conceptual, and iconographic inputs generated to fool the model into making false or absurd classifications. Zhang et al.~\cite{zhang2022towards} propose a multimodal attack method that collectively carries out the attacks on the image modality and the text modality. Hintersdorf et al.~\cite{hintersdorf2022clipping} introduce to assess privacy for multi-modal models and reveal whether an individual was included in the training data by querying the model with images of the same person. Jia et al.~\cite{jia2022badencoder}  injects backdoors into a pre-trained image encoder.% such that the target classifiers built based on the backdoored image encoder for different downstream tasks simultaneously inherit the backdoor behavior.

In this work, we utilize the VL models to establish the attack frameworks.  Previous works, however, either focus on one modality or treat the two modalities separately, which constrains transferring to target networks or other datasets.

\section{Proposed Method}
%In this paper, we explore the generation of adversarial samples that can fool downstream black-box networks. We begin by providing an overview of the entire framework, followed by a description of the attack-and-defense framework in detail.
\subsection{Framework Overview}
In the proposed framework, we utilize the generator-orientated adversarial attack method, which integrates CLIP to enable the transferability.
That is, from a training distribution of images, we train the generator $\mathcal{G}$ to generate universal perturbations applied on the input clean image $x_i$. The corresponding adversarial sample ${x}_i'$ can be thus obtained as:
\begin{equation}
    {x}_i' = \min\big(x_i+\epsilon,\max(\mathcal{G}(x_i),x_i-\epsilon)\big),
    \label{eq:bound}
\end{equation}
where $\epsilon$ is the predetermined maximum perturbation on the input. Each applied perturbations are bounded with $[-\epsilon,+\epsilon]$, in the rest of the paper, we simplify this process and use $\mathcal{G}(x_i)$ to denote the bounded adversarial sample.

Denote the groundtruth label of the clean image $x_i$ as $y_{\text{true}}$, then the attack goal is to obtain the universal perturbations enabling cross-architecture and cross-dataset transferring.
The goal of attacking a group of black-box  models $\mathcal{M}=\{\mathcal{M}^0, \mathcal{M}^1,...,\mathcal{M}^{P-1}\}$ pre-trained on various datasets $\mathcal{D}=\{\mathcal{D}^0, \mathcal{D}^1,...,\mathcal{D}^{P-1}\}$ is to maximize:
\begin{equation}
 \sum\mathbf{1}_{y'\neq y_{\text{true}}}| y'\! \leftarrow \mathcal{M}^p\big[\mathcal{G}(x_i)\big], \quad x_i\in \mathcal{D}^p, \mathcal{M}^p\in \mathcal{M}
\end{equation}
where $\mathbf{1}$ is the indicator function.

 To achieve it, we utilize CLIP as the surrogate model to train the powerful generator $\mathcal{G}$, as is shown in Fig.~\ref{fig:framework}. 
%Considering that CLIP is a powerful model pre-trained on a diverse range of internet-scale image-text pairs (400 million images and their associated captions, alt-text, and web pages), it is capable of learning the relationship between different modalities. Thus CLIP shows that we can use this knowledge to make predictions even for previously unseen combinations of images and text.
Considering that CLIP is a powerful model capable of learning the relationship between different modalities, 
we utilize CLIP as the surrogate model to generate the perturbations.
Denote the textual encoder to be $E_i$ and the visual encoder to be $E_t$, then CLIP is capable of zero-shot predictions by matching the most similar textual embedding as:
 \begin{equation}
\begin{split}
p(y|x_i)&=Clip\big(x_i,X_t\big)\\
&=\frac{\exp[\text{sim}\big(E_i(x_i), E_t(x_{t}^y)\big)/\tau]}{\sum_{c=0}^{C-1}\exp[\text{sim}\big(E_i(x_i), E_t(x_{t}^c)\big)/\tau]},
\end{split}
 \end{equation}
 where $\text{sim}(\cdot)$ is the cosine similarity, $\tau$ is the temperature parameter.  And $X_t=\{x_t^0,x_t^1,...,x_t^{C-1}\}$ is text modality input that generates for each classification label $c$ ($c\in \{0,1,...,C-1\}$).  
 
As the CLIP takes both two modalities as input to make predictions as $y'=\arg\max_y p(y|x_i)$, we construct the attack and defense into one framework on these two modalities.
 Specifically, producing attack with CLIP can be treated as a kind of adversarial training process:
 \begin{equation}
 \begin{split}
&\max_{\mathcal{P}}\min_{\mathcal{G}} V(\mathcal{P},\mathcal{G})\\
&= \mathbb{E}_{x_i\sim p_{X_i}}\text{sim}\big[Clip\big(\mathcal{G}(x_i),\mathcal{P}(X_t)\big), Clip\big(x_i,\mathcal{P}(X_t)\big)\big]\\
&+ \mathbb{E}_{x_t\sim p_{X_t}}\text{sim}\big[Clip\big(\mathcal{G}(X_i),x_t\big), Clip\big(\mathcal{G}(X_i),\mathcal{P}(x_t)\big)\big],\\
 \end{split}
 \label{eq:adv}
 \end{equation}
where $\mathcal{P}(\cdot)$ is the prompt tuning function for the textual input, the details of which are in Sec.~\ref{sec:defense}
Similar as the GAN training, the optimization of Eq.~\ref{eq:adv} is to iteratively update $\mathcal{G}$ and $\mathcal{P}$.
The whole process is, 
\begin{itemize}
\item for optimizing $\mathcal{G}$ to \textbf{generate the perturbations} on the clean image $x_i$, we minimize the output similarity between the clean input $x_i$ and the adversarial input $x_i'$, which is in \textit{`Visual Attack with Semantic Perturbation'}; 
\item for optimizing $\mathcal{P}$ to \textbf{defend the attack} on the image modality input, we tune the prompt as $\mathcal{P}(x_t)$ to match the image-text embedding again, which is in \textit{`Textual Defense with Prompt Updating'};
\item with the \textbf{iterative training} of $\mathcal{G}$ and $\mathcal{P}$, we obtain the final generative perturbation network $\mathcal{G}$, the whole algorithm is given in the supplementary.
\end{itemize}

\subsection{Visual Attack with Semantic Perturbation}
Recall that the visual attack $x_i'$ generated from $\mathcal{G}$ is bounded with $\epsilon$ (in Eq.~\ref{eq:bound}), it is assumed to be assigned with the wrong prediction by the target network $\mathcal{M}$.
Considering the fact that $\mathcal{M}$ keeps as the black-box, we turn to attack on the feature embedding space of the CLIP model.

The CLIP model's pre-trained image encoder $E_i$ is a powerful feature extractor that has high transferability. To ensure the transferability of adversarial samples, we aim to maximize the distance between the feature representation of the adversarial input $x_i'$, denoted as $E_i(x_i')$, and the feature representation of the clean input $x_i$, denoted as $E_i(x_i)$. This is achieved by minimizing the loss function $\ell_{feat}$, which is calculated as: 
\begin{equation}
\begin{split}
\ell_{feat} = -&\|\mathcal{F}_i-\mathcal{F}_i'\|^2\\
\mathcal{F}_i =\frac{E_i(x_i)}{\|E_i(x_i)\|},& \mathcal{F}_i'=\frac{E_i(x_i')}{\|E_i(x_i')\|},
\end{split}
\label{eq:feat}
\end{equation}
where $\ell_{feat}$ is calculated based on the MSE loss.

Other than maximizing the features similarity with the perturbed and the clean input, an extra triplet loss is applied to ensure that the perturbed features ${E_i(x_i')}$  fool the down-stream networks with a wrong prediction. 
To calculate the triplet loss, the original textual embedding should be pre-computed as $\mathcal{F}_t^c = {E_i(x_t^c)}/{\|E_i(x_t^c)\|}$ ($c\in \{0,1,...,C-1\}$).
Each $x_t^c$ is composed of a prompt template and a label object, i.e. `dog' + `A clean photo of  \{\}'. Thus, for each label $c\in\{0,1,...,C-1\}$, we pre-compute the corresponding textual embeddings as $\mathcal{F}_t=\{\mathcal{F}_t^0,\mathcal{F}_t^1,...,\mathcal{F}_t^{C-1}\}$.
In this way, for each clean image $x_i$ with the groudtruth label $y_{\text{true}}$, we use the the triplet loss ~\cite{aichgama} $\ell_{tri}$ to mislead the match of the features from the two modalities.
Specifically, $\ell_{tri}$ is calculated as:
\begin{equation}
\begin{split}
    \ell_{tri} =  \|\mathcal{F}_i'-\mathcal{F}_t^{y'}\|^2&+\max\big(0, \alpha-\|\mathcal{F}_i'-\mathcal{F}_t^{y_{\text{true}}}\|^2\big),\\
    where \quad y'&=\arg\min_c\text{sim}(\mathcal{F}_i, \mathcal{F}_t^c),
\end{split}
\label{eq:tri}
\end{equation}
where $\alpha$ is the margin of the triplet loss. $\mathcal{F}_t^{y'}$ is the textual embedding that is least similar to that of the clean input.
 This triplet loss forces the perturbed features away from its groundtruth textual embedding $\mathcal{F}_t^{y_{\text{true}}}$ while minimizing the distance with the textual embedding $\mathcal{F}_t^{y'}$ that is originally least related to the clean image features $\mathcal{F}_i$.

Finally, following the previous adversarial attack methods, we utilize an extra classification loss:
\begin{equation}
    \ell_{cls}=\frac{1}{\sigma+H_{CE}\big(Clip(x_i', X_t), y\big)},
\label{eq:cls}
\end{equation}
where we set $\sigma= 0.1$ to prevent gradient explosion. $H_{CE}(\cdot)$ is the standard cross-entropy loss, and $Clip(\cdot)$ is the output probabilities after softmax.

Thus, the final learning objective for visual attack is:
\begin{equation}
    \arg\min_{\mathcal{G}}  \ell_{feat} + \ell_{tri} + \ell_{cls},
\end{equation}
with which, the optimized $\mathcal{G}$ is capable of generating perturbations that could fool the textual input in the form of $X_t$ and have a certain degree of transferablity.

\label{sec:attack}
\subsection{Textual Defense with Prompt Updating}
\label{sec:defense}
For a clean image $x_i$, its feature embedding can be denoted as $\mathcal{F}_i$.
For each label $c \in\{0,1,...,C-1\}$, the text input for each label can be organized by $m+1$ text tokens as: $x_t^c=[<\text{CLASS}(c)>, v_1, v_2,...,v_m]$.
Then the textual embedding for each label $c$ is calculated as $\mathcal{F}_t^c \leftarrow E_t(x_t^c)$.
And the CLIP model tends to output the probabilities for each label as:
\begin{equation}
\begin{split}
p(y|x_i, X_t)&=Clip(x_i, X_t) \leftarrow softmax(\mathcal{F}_i*\mathcal{F}_t)\\
& where\quad   X_t = X_l + X_p.
\end{split}
\end{equation}
Here, we separate the text input $X_t$ into the fixed label token $X_l=<\text{CLASS}(c)>$ and the dynamic prompt token $X_p=\{v_1,v_2,...,v_m\}$.
Previous work on prompt learning~\cite{zhou2022learning} has indicated that the position of the label token wouldn't make big difference on the final results, we intuitively put the label token at the beginning.

\begin{table*}[t]
\centering
\scriptsize
\caption{Ablation study on attacking CLIP. The experiments are conducted on both CIFAR-10 and ImgaeNet datasets. The attacks are obtained by the CLIP model and are tested with the CLIP model and a target pre-trained ResNet-50. We marked the
best overall performance as bold (best viewed in color).}
\vspace{-1em}
\begin{tabular}{ccp{6mm}<{\centering}p{6mm}<{\centering}p{6mm}<{\centering}p{6mm}<{\centering}p{9mm}<{\centering}p{9mm}<{\centering}p{6mm}<{\centering}p{6mm}<{\centering}p{6mm}<{\centering}p{6mm}<{\centering}p{9mm}<{\centering}p{9mm}<{\centering}}
\toprule
\multirow{2}{*}{\textbf{Method}} & \multirow{2}{*}{\textbf{Surrogate}} & \multicolumn{6}{c}{\textbf{CIFAR-10 (Train/ Val)}} & \multicolumn{6}{c}{\textbf{ImageNet (Train/ Val)}} \\\cmidrule(r){3-8} \cmidrule(r){9-14}
 &       &  \multicolumn{2}{c}{ \textbf{CLIP}}   & \multicolumn{2}{c}{\textbf{ResNet-50 }}& \multicolumn{2}{c}{\textbf{Overall}}   &  \multicolumn{2}{c}{\textbf{ CLIP}}   & \multicolumn{2}{c}{\textbf{ResNet-50}}& \multicolumn{2}{c}{\textbf{Overall}} \\ \hline
 Clean& - & 88.3  & 88.5 &100.0&94.6& 94.2 &91.6 &59.1&59.0 & 75.9 & 76.5&67.5 & 67.8 \\
White-box Attack    & ResNet-50   & 64.2 & 64.2 & 13.7&  13.4 & 39.0\down{-55.2}& 38.8\down{-52.8}&42.0&41.9 & 5.7 &6.0 &23.9\down{-43.6} & 24.0\down{-43.8}    \\ 
w/o $\ell_{feat}$ & CLIP & 12.1 &12.4 &53.2 &54.7 &32.7\down{-61.5} & 33.6\down{-58.0}&9.6 &9.7 & 55.4&54.7&32.5\down{-35.0} &32.2\down{-35.6} \\ 
w/o $\ell_{tri}$  & CLIP & 11.9 &11.9 & 51.0 & 52.0 & 31.5\down{-62.7} & 32.0\down{-59.6}& 9.7 & 8.9 & 54.2 & 55.7 & 32.0\down{-35.5} & 32.3\down{-35.5}\\ 
w/o $\ell_{cls}$  & CLIP & 10.1&10.6 & 52.8&53.1&31.5\down{-62.7} & 31.9\down{-59.7}&10.6 & 12.2 & 59.2 &59.8 & 34.9\down{-32.6} & 36.0\down{-31.8} \\ 
Visual Attack w/o Iter  &  CLIP & 8.1 & 8.9 & 54.5 & 55.2 & 31.3\down{-62.9}& 32.1\down{-59.5}&8.8& 9.7 & 53.9 & 54.3 & 31.4\down{-36.1} & 32.0\down{-35.8}\\ \midrule
Random Prompt & CLIP & 8.4 & 8.6 & 55.2 & 56.4 &31.8\down{-62.4} & 32.5\down{-59.1}&8.3 & 9.1 & 41.3 & 39.7 & 24.8\down{-42.7} & 24.4\down{-43.4}\\
Ours(Full)&  CLIP & 7.9& 7.2 & 41.3&41.8 & \textbf{24.6\down{-69.6}} & \textbf{24.5\down{-67.1}}& 7.5 &7.8 & 25.0& 25.3 &  \textbf{16.3\down{-51.2}}& \textbf{16.6\down{-51.2}}\\
 \bottomrule
\end{tabular}
\label{tab:abl}
\end{table*}

Supposing that based on the current text input $X_t$ the CLIP model makes the right prediction on the clean input $x_i$ with the groundtruth label $y_{\text{true}}$, and the attack generator has successfully attacked it by predicting it as a wrong label $y'$.
The attack (A) and the defense (D) process could be formulated as:
\begin{equation}
\begin{split}
&[A]:\ \arg\max_yp\big(y|{\mathcal{G}(x_i)},X_t\big) \neq \arg\max_yp\big(y|x_i,X_t\big). \\
&[D]:\  \arg\max_yp\big(y|\mathcal{G}(x_i),\mathcal{P}{(X_t)}\big) = \arg\max_yp\big(y|x_i,X_t\big),
\end{split}
\end{equation}
where in [A], we learn the generator $\mathcal{G}$ for generating adversarial perturbations and in [D], we update the text input $X_t$ to $X_t '$ with the prompt tuning function $\mathcal{P}$ to guide the right prediction on CLIP again.

During the prompt tuning, we fix the label tokens $X_l$, and only update the prompt template $X_p '$ by maximizing the semantic similarity comparing with the former visual embeddings:
\begin{equation}
\begin{split}
X_t' =\Big\{ x_t^c|& c\in\{0,1,..,C-1\},\\
&{x_t^c}' \leftarrow\arg\min_{x_t^{y_{\textbf{true}}}} \text{Sim} [E_i(x_i'), E_t(x_t^{y_{\textbf{true}}})]
    \Big\}
\end{split}
\end{equation}
where $X_l^c$ is the $c$-th label token. However, due to the fact that it is impractical  to directly learn each optimal text token, we turn to modify the  Probability Weighted Word Saliency~\cite{Ren2019GeneratingNL} method  for each word token by randomly replacing each word token $v_n\subset X_p$ as:
\begin{equation}
\begin{split}
    S(v_n) &= \max\big[p(y'|x', X_p) - p(y'|x', X_p^n),0\big],\\
    where \quad
   & X_p^n = \{v_1,...v_{n-1},<\text{MASK}>,...,v_m\},
\end{split}
\label{eq:saliency}
\end{equation}
where we mask each word token to calculate the saliency score that fools the model to make the  wrong prediction $y'$ on the adversarial sample $x'$. 
We set the threshold value to be $\rho$, meaning that only the word tokens $X_{\text{update}}=\{v_n|S(v_n)>\rho, 1\le n\le m\}$ are set to be updated.

Thus, we update each word token in $X_{\text{update}}$ from a set of candidates, the updating process is formulated as:
\begin{equation}
\begin{split}
v_n*=\arg\max_{v_n'} & p(y_{\text{true}}|x', X_p(v_n')\! -\! p(y_{\text{true}}|x', X_p^n)), \\
X_p(v_n') &= \{v_1,...v_{n-1},\mathbf{v_n'},...,v_m\},\\
 v_n \in & X_{\text{update}}\quad and \quad v_n'\in \Gamma(v_n),
\end{split}
\label{eq:promptune}
\end{equation}
where $\Gamma(v_n)$ is the candidate word set generated by GPT-2~\cite{radford2019language}. And each candidate word token is updated to correct the semantic consistency again by ensuring most of the perturbed samples are re-matched with its groundtruth related word embedding.

As a whole, the prompt tuning function can be donated as:
$\mathcal{P}(X_t)=X_l+X_p(\cup_n v_n*)$.
And the full algorithm can be found in the supplementary.

\section{Experiments}
In our experiments, we evaluated the attack performance of our proposed framework on several publicly available benchmark datasets. As our framework aims to generate highly transferable attacks, we focused on evaluating the transferability of the attacks in cross-dataset and cross-architecture settings.
More experimental settings and results can be found in the supplementary.

\subsection{Settings}
\textbf{Datasets.}
Followed previous work~\cite{Hayes2017LearningUA}, We evaluate attacks using two popular datasets in adversarial examples research, which are the CIFAR-10 dataset~\cite{Krizhevsky2009LearningML} and the ImageNet dataset~\cite{Russakovsky2014ImageNetLS}. 
For testing the cross-dataset transferability, we follow previous works~\cite{naseer2019cross,salzmann2021learning} and use the Comics and Paintings~\cite{painter} or ChestX datasets as source domain, and evaluate on the randomly selected 5000 images from the ImageNet.
%More experiments on other datasets are included in the supplementary.
%The CIFAR-10 dataset consists of 60,000, 32×32 RGB images of different objects in ten classes, which are airplane, automobile, bird, cat, deer, dog, frog, horse, ship, truck. 
%For the ImageNet dataset that contains 1,000 classes, we use its validate set, which is split into 50,000 images for training and 10,000 images for validation.

\textbf{Implemental Details. }
We used PyTorch framework for the implementation.
In the normal setting of using the pre-trained CLIP as the surrogate model, we choose the `ViT/32' as backbone.
As for the generator, we choose to use the ResNet backbone, and set the learning rate to be 0.0001 with Adam optimizer.
All images are scaled to $224\times 224$ to train the generator.
For the $\ell_{\infty}$ bound, we set $\epsilon = 0.04$.
A total of 10 iterations (we set the $NUM_G$ to be 2) are used to train the whole network,
which costs about 8 hours on one NVIDIA GeForce RTX 3090 GPUs.

\textbf{Inference Metrics.}
We evaluate the performance of the generated attack by the mean accuracy of the classifier, which is better with lower values.
In addition, as we aim at generating the high-transferable attacks, which is evaluated with various down-stream networks and datasets.
We get the overall accuracy be the group-wise average, where the architectures with similar architectures are calculated once, i.e. we average the accuracies on ResNet-based architecture.

\subsection{Experimental Results}

%\subsubsection{Overall Performance}

\textbf{Ablation study on the proposed framework.}
We conduct the ablation study on the proposed framework in Table~\ref{tab:abl}, where the methods listed for comparison are:
\begin{itemize}
    \item Clean: clean input without any attack;
    \item White-box Attack: we generate the attack directly on the target network ResNet-50, which serives as the upperbound;
    \item w/o $\ell_{feat}$/$\ell_{tri}$/$\ell_{cls}$: the visual attack optimized without the loss item $\ell_{feat}$/$\ell_{tri}$/$\ell_{cls}$;
    \item Visual Attack w/o Iter: the visual attack with semantic consistency optimized with loss $\ell_{feat} + \ell_{tri}+\ell_{cls}$;
    \item Random Prompt: we use GPT-2 to randomly generate $X_p$ in each iteration.
\end{itemize}
As can be observed from the table:
(1) The proposed method achieves the best attack performance in  `Overall', which fools the classifier by decreasing the accuracies more than 60\% on CIFAR10 and more than 50\% on ImageNet.
(2) Only applying the visual attack on CLIP (`Visual Attack w/o Iter') can successfully attack the CLIP
Model, but transfers bad on the down stream network.
(3) Randomly update the prompt templates (`Random Prompt') can't improve the attack's transferablity a lot, indicating the effectiveness of the proposed prompt tuning $\mathcal{P}(\cdot)$.
\begin{figure}[t]
\centering
\includegraphics[scale = 0.42]{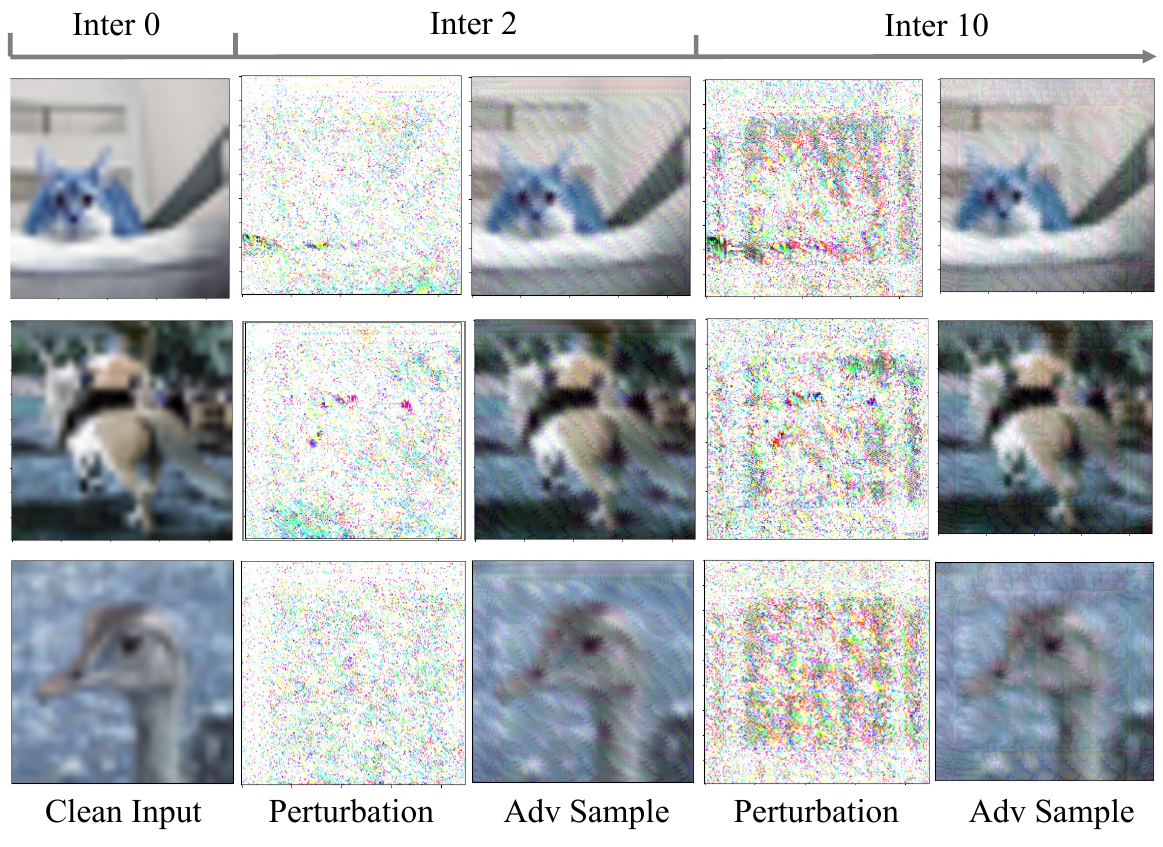}
\caption{ The generated perturbations and adversarial samples on CIFAR-10 dataset. The visualizations are depicted on the 2-nd iteration and 10-th iteration. }
\label{fig:adv}
\end{figure}

\begin{figure}[t]
\centering
\includegraphics[scale = 0.45]{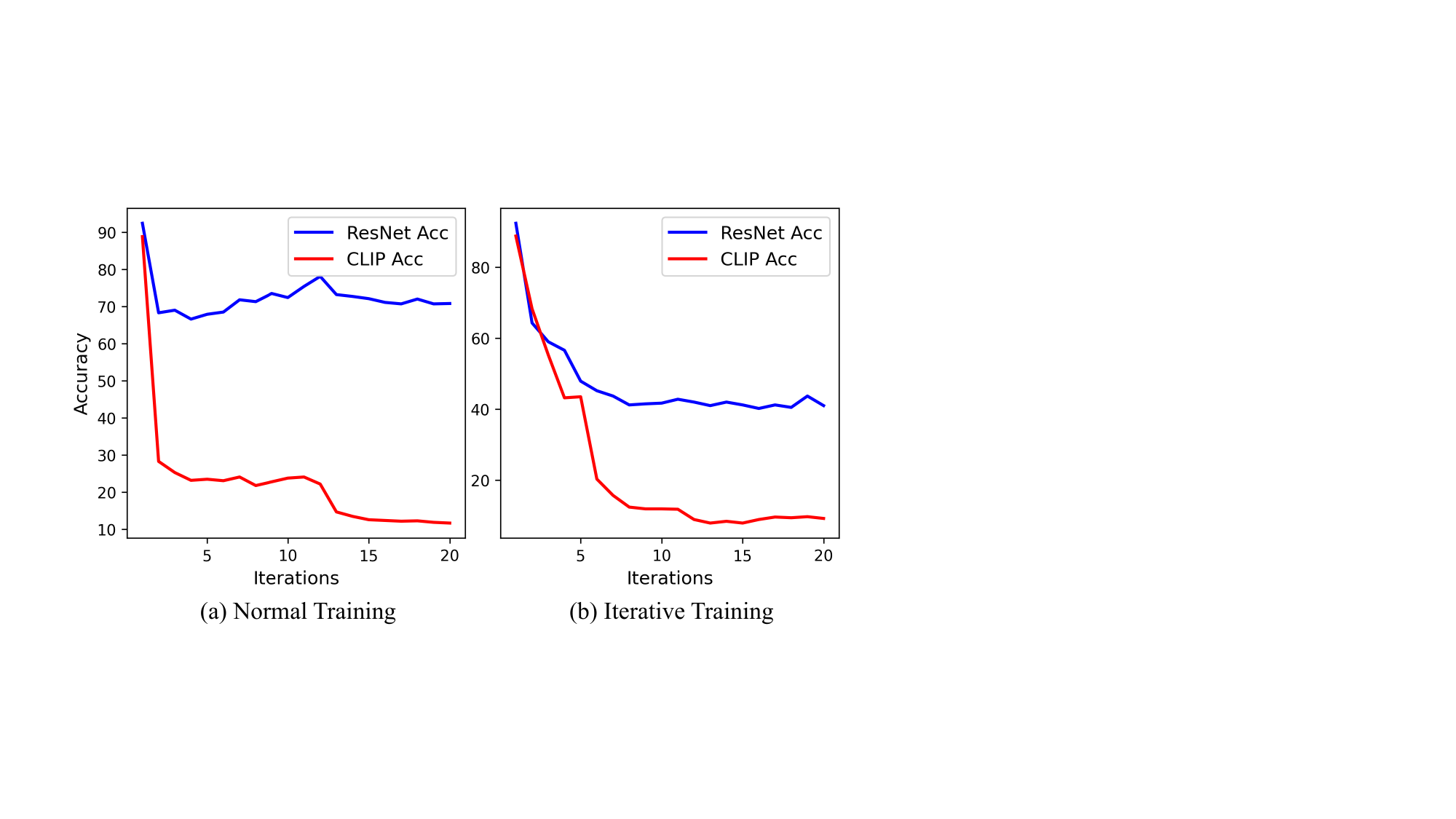}
\caption{ The attack's performance from the generative perturbation network in each iteration training. The experiments are conducted on CIFAR10 dataset. }
\vspace{-1em}
\label{fig:iter}
\end{figure}

\begin{figure}[t]
\centering
\includegraphics[scale = 0.67]{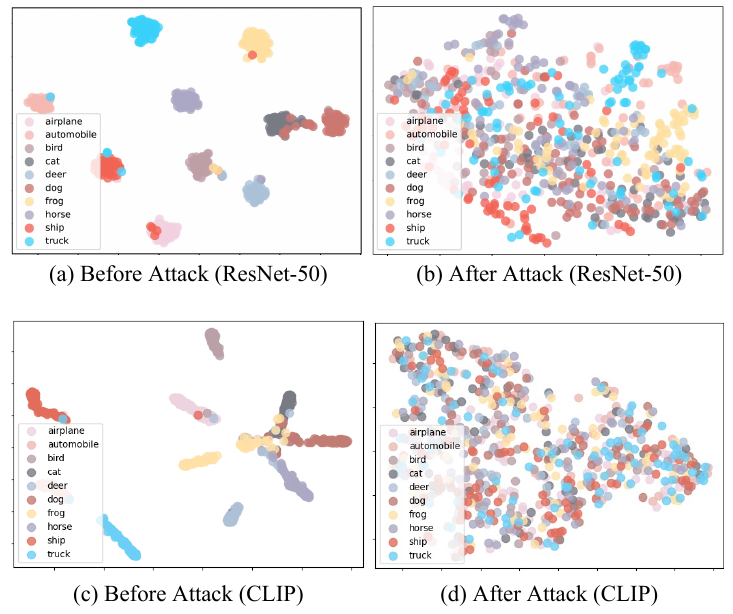}
\caption{ Embedding visualization. We do the TSNE visualization on the features with the clean/adversarial images as the input (`Before/After Attack').}
%Sub figures (a) and (b) are visualized based on the intermediate features of the pre-trained ResNet-50. Sub figures (c) and (d) are visualized based on the intermediate features of the pre-trained CLIP model.}
\label{fig:tsne}
\vspace{-1em}
\end{figure}
\begin{table*}[ht]
\centering
\scriptsize
\caption{Comparative results on the  cross-architectures transferability. We evaluate the classification accuracy (lower is better) and show the results on both CIFAR-10 and ImageNet datasets. Our proposed method achieves the best in the `overall' metric (best viewed in color).}
\begin{tabular}{cccccccccccc}
\toprule
\multirow{2}{*}{\textbf{Method}} & \multirow{2}{*}{\textbf{Dataset}} & \multicolumn{9}{c}{\textbf{Transfer to Target Networks}} & \multirow{2}{*}{\textbf{Overall}} \\ \cmidrule{3-11}
                        &                          & CLIP & ResNet18   &   ResNet34  &  ResNet50   &  VGG11  &   VGG19 &  ShuffleNetv2 &  MobileNetv2 & SimpleViT  &                          \\ \midrule
\multirow{2}{*}{Clean} & ImageNet  &  59.0 & 70.3  & 73.6 & 76.5& 69.2 & 72.5 &69.8 & 72.1 &80.9&71.0                     \\
             &   CIFAR-10          &  88.5  & 94.6 & 94.7 & 94.9 & 92.2 & 93.1 & 90.2 & 93.9 & 81.7  & 89.9                 \\ \midrule 
\multirow{2}{*}{UAN-Res} & ImageNet & 41.9 & 18.5 & 20.1 & 6.0 & 16.4 &  15.6 &  32.4 & 11.3  & 60.5  &  31.0\down{-40.0}                   \\
             &   CIFAR-10 &  64.2  &  19.6& 23.9 & 13.4 &71.7 & 38.7 &58.2 & 15.3 &31.7 & 41.4\down{-48.5}              \\ \midrule 
\multirow{2}{*}{UAN-Clip} & ImageNet &  17.3 & 29.8 &  35.3 & 37.3 & 21.8 & 22.5 & 34.1& 28.5& 53.9 & 31.8\down{-39.2}
                    \\
             &   CIFAR-10  & 10.6 &   50.2 & 52.9  & 54.5 & 70.8 & 40.5  & 55.1 &  30.9 &  25.5 &  37.5 \down{-52.4}  \\\midrule 
\multirow{2}{*}{BIA-VGG} & ImageNet    & 45.9 &  23.6 &   23.7 & 25.4 &  9.0  & 3.6 &  30.4 & 20.6 & 62.2  &  32.8\down{-38.2}                 \\
             &   CIFAR-10          & - &  - &  - & -  &  - &  - & -& - &  -&-                         \\  \midrule
\multirow{2}{*}{Ours-Clip} & ImageNet& 7.8&  17.3 & 22.8 & 25.3&15.6 &  19.9 & 23.4 & 25.7 & 44.8 &  \textbf{23.3\down{-47.7}}\\
             &   CIFAR-10  &7.2 & 38.5 & 40.2 & 41.8 &  64.5 & 38.9 & 45.2 & 20.0 &  20.6 & \textbf{ 30.5\down{-59.4}}  \\  
            \bottomrule
            
\end{tabular}
\label{tab:crossarch}
\end{table*}

\textbf{Visualization the generated adversarial samples.}
We show the generated adversarial samples in Fig.~\ref{fig:adv}.
In the figure, we sample the perturbation generator at the 2-nd iteration and the 10-th, respectively.
It can be observed from the figure that the perturbations generated from the pre-trained CLIP model tend like some regular patterns. And this perturbation patterns are strengthened with the iterative training.
It is worth noting that when generating the targeted perturbations, we find more interesting patterns, which could be found in the supplementary.

\textbf{The performance during the iterative training.}
The main idea of the proposed framework is to utilize the iterative training strategy. Here, we depict the generated attack's performance while the iterative training in Fig.~\ref{fig:iter}, where we compare the performance on normal training the generator $\mathcal{G}$ (a) and the proposed iteratively training the generator (b).
As can be seen in the figure, we evaluate the attack capacity on both the surrogate network to train the generator (CLIP) and the target network (ResNet50).
Thus, the observations are:
(1) In the normal training scheme, the generator convergences faster, but finally the attack success rate is lower than the proposed iterative training.
(2) In the normal training scheme, the attack's transferablity performance on the ResNet improves at first, but fails in the afterward iterations. While in our proposed framework, the transferablity improves stably, which is mainly due to the updating on the other modality.
(3) About 10-iteration training would optimize an optimal perturbation generator, thus, we set the total iteration number to be 10 in the rest of experiments.

\textbf{Analysis on the embedding visualization.}
We visualize the embedding features in Fig.~\ref{fig:tsne}. The visualization includes the feature space before and after attack and is conducted on the CIFAR-10 dataset, where the following observations can be made:
(1) The features belonging to the same category are grouped together, making it easy for the classifier to recognize them. However, the feature space after attack is mixed together, which fools the classifier.
(2) Comparing the visualized features after attack of CLIP (d) and ResNet-50 (c), the adversarial features generated by our work are mixed together more evenly. This makes it much more difficult to defend against and indicates a higher attack capability.
\subsubsection{Transferability Evaluation}

We have compared with other methods on the cross-architecture, cross-dataset (domain) and cross-task transferability.

\textbf{Evaluate the cross-architecture transferability}
We form a set of pre-trained networks in various architectures, which are divided into 5 groups: (1) CLIP, (2) ResNet18, ResNet34, ResNet50, (3) VGG11, VGG19, (4) ShuffleNetv2, MobileNetv2 and (5) SimpleViT. Based on these grouping strategy, we calculate the overall accuracy by group-average accuracy. The experimental results are compared in Table~\ref{tab:crossarch}.
We have compared the proposed method with the other generator-oriented  methods, which are UAN~\cite{Hayes2017LearningUA} (`UAN-Res'), modified UAN that uses CLIP as the surrogate model (`UAN-CLIP') and BIA~\cite{zhang2022towards}.% BIA uses VGG-16 as surrogate model and since BIA is proposed on ImageNet, we only show its performance trained on ImageNet.
From the table, we observe that:
(1) When transferring the generated attack to the target networks, the attacks perform better between the networks in similar architecture as the surrogate model;
(2) We propose to generate the attacks with high transferability, which decrease the overall accuracies most on both ImageNet and CIFAR-10 datasets;
(3) The other attack methods (`UAN' and `BIA') perform unevenly according to the target networks, which could be easily defended by the ensemble attack; while \textbf{the proposed mutual-modality attack is stable whatever the target networks}, making it more difficult to defend. The corresponding experiment against the ensemble defense is included in the supplementary.

\begin{table}[t]
\scriptsize
\caption{Extreme cross-domain (dataset) transferability analysis evaluated by attack success rate.}
\vspace{-1mm}
\centering
\begin{tabular}{p{7mm}<{\centering}p{7mm}<{\centering}p{12mm}<{\centering}p{11mm}<{\centering}p{11mm}<{\centering}p{11mm}<{\centering}}
\toprule
\multirow{2}{*}{\textbf{Methods}} & \multirow{2}{*}{\textbf{Datasets}}   & ResNet-152 & CLIP & SimpleViT & Overall  \\ \cmidrule{3-6} 
              &    & \multicolumn{4}{c}{Curr. / Curr. + Ours}                      \\ \midrule
\multirow{3}{*}{GAP }& Cosmics &50.3/51.2\down{+0.9} &20.3/30.8\down{+10.5} & 27.6/35.7\down{+8.1} &32.7/39.2\down{+6.5} \\
& Paintings &52.9/53.0\down{+0.1}&36.7/37.7\down{+1.0} &37.8/46.9\down{+9.1} &42.5/45.9\down{+3.4}\\
& ChestX &29.2/32.8\down{+3.6}&19.8/36.4\down{+16.6}&19.4/38.7\down{+19.3} & 22.8/36.0\down{+13.2}\\ \midrule
\multirow{3}{*}{CDA} & Cosmics &38.8/39.6\down{+10.8}&36.8/46.6\down{+9.8}&37.6/42.6\down{+5.0} &37.7/42.9\down{+5.2}\\
& Paintings &41.7/41.8\down{+0.1}&38.6/43.5\down{+4.9}&39.0/46.2\down{+7.2} & 39.8/43.9\down{+4.1}\\
& ChestX & 23.7/30.3\down{+6.6}&16.7/29.3\down{+12.6}&19.2/27.9\down{+8.7} & 19.9/29.2\down{+9.3}\\ \midrule
\multirow{3}{*}{LTAP} & Cosmics &55.2/56.5\down{+1.3}&43.6/45.5\down{+1.9}&48.4/54.0\down{+5.6} &49.1/52.0\down{+2.9}\\
& Paintings &59.9/60.7\down{+0.8}&44.8/53.5\down{+8.7}&48.6/48.7\down{+0.1}&51.1/54.3\down{+3.2} \\
& ChestX &49.5/50.3\down{+0.8}&28.9/34.2\down{+5.3}&21.9/24.6\down{+2.7} & 33.4/36.4\down{+3.0}\\
\bottomrule
\end{tabular}
\label{tab:crossdata}
\end{table}

\iffalse
\begin{table}[t]
\scriptsize
\caption{Comparison results on the attacks' transferability on cross-datasets. We evaluate the classification accuracy and show the results on both ImageNet $\rightarrow$ CIFAR-10 and CIFAR-10 $\rightarrow$ ImageNet.}
\centering
\begin{tabular}{cccccc}
\toprule
\multirow{2}{*}{\textbf{Method}} & \multirow{2}{*}{\textbf{ImageNet}} & \multicolumn{4}{c}{\textbf{ImageNet} $\rightarrow$ \textbf{CIFAR-10}}  \\ \cmidrule{3-6} 
                        &                           & ResNet-50 & CLIP & SimpleViT & Overall \\ \hline
Clean  & -  &   94.9 &   88.5  &   81.7  & 88.4    \\
UAN-Res   & 6.0  &  74.0  &   69.2 &  62.1  &68.4\down{-20.0}       \\
UAN-Clip   &   17.3 &  84.3   &  70.5 &70.4   & 75.1\down{-13.3}     \\
BIA-VGG & 3.6 & 71.1 & 70.8 & 72.3& 71.4\down{-17.0}\\
Ours-Clip & 7.8 &  75.6  &  58.8  & 60.1  &  \textbf{64.8\down{-23.6}}  \\ \midrule\midrule
\multirow{2}{*}{\textbf{Method}} & \multirow{2}{*}{\textbf{CIFAR-10}} & \multicolumn{4}{c}{\textbf{CIFAR-10} $\rightarrow$ \textbf{ImageNet}  }  \\ \cmidrule{3-6} 
   &      & ResNet-50 & CLIP & SimpleViT & Overall \\ \hline
Clean   &   -   & 76.5  &  59.0 &  80.9 &  72.1       \\
UAN-Res  &    13.4  & 69.9  &  56.1  &  80.5  & 68.8\down{-3.3}     \\
UAN-Clip &  10.6   & 70.7   & 52.6    &  79.8   &  67.7\down{-4.4}     \\
Ours-Clip  & 7.2    &    68.4    &   45.9   &  64.3   & \textbf{59.5\down{-12.6}}     \\ 
\bottomrule
\end{tabular}
\label{tab:crossdata}
\end{table}
\fi

\textbf{Evaluate the Cross-dataset Transferability.}
We evaluate the cross-data transferability by training the generator on the source dataset and test on the target dataset.
Following the previous setting, we train the generator on Comics/Paintings/ChestX datasets with ChestXNet as the discriminator and evaluate the attack performance on ImageNet.

As we propose a plug-and-play method, we test the effectiveness of our method by integrating it into the existing methods, including: GAP~\cite{poursaeed2018generative}, CDA~\cite{naseer2019cross} and  LTAP~\cite{salzmann2021learning}.
As can be observed from the table:
(1) Enabling the cross-dataset transferability of the attacks is much more difficult than the cross-architecture one, and our method also shows satisfying results;
(2) \textbf{Our method (` Curr. + Ours') improves the cross-dataset attack success rates when integrated into the current methods} especially on the cases with CLIP and SimpleViT as target networks.

\textbf{Evaluate the Cross-task Transferability.}
Following previous work~\cite{zhang2022beyond} we conduct the cross-task tranferability evaluation in Table~\ref{tab:crosstask}.
We integrate the proposed framework into GAMA~\cite{aichgama} (`GAMA+Ours').
We train the generator with Pascal-VOC dataset~\cite{Everingham2010ThePV}, and then test on the ImageNet classification task.
As cab be observed from the figure, our proposed framework could be integrated into any adversarial attack framework, which could further enhance the attack transferability.
\begin{table}[t]
\centering
\scriptsize
\caption{The cross-task transferability evaluation.}
\vspace{-2mm}
\begin{tabular}{ccccc}
\toprule
\textbf{Method}&\textbf{VGG-16}&\textbf{ResNet-50}&\textbf{SimpleViT} &\textbf{Overall} \\\midrule
Clean & 51.3 & 69.9 & 54.7 & 58.6\\
GAMA&  \textbf{3.1\down{-48.2}}&22.3\down{-47.6} & 38.3\down{-16.4} &21.2\down{-37.4}\\
GAMA+Ours &  3.4\down{-47.9} & \textbf{20.4\down{-49.5}}& \textbf{35.9\down{-18.8}} & \textbf{19.9\down{-38.7}}\\
\bottomrule
\end{tabular}
\label{tab:crosstask}
\vspace{-4mm}
\end{table}

\section{Conclusion}
Overall, our proposed approach demonstrates promising results in improving the transferability and stability of adversarial attacks by generating perturbations in the semantic feature embedding space using the pre-trained CLIP model. By optimizing the attack iteratively from both image and text modalities, our method achieves improved transferability across different architectures and datasets, as demonstrated in our experiments on several benchmark datasets.
We believe that our work provides a valuable contribution to the field of adversarial attacks and could have important implications for improving the security and reliability of machine learning systems in real-world applications.
%Our approach offers a practical and efficient solution for improving the robustness of adversarial attacks against machine learning systems, and could be readily deployed as a plug-and-play solution in various settings.
\section*{Acknowledgements}
This research is supported by the Ministry of Education, Singapore, under its Academic Research Fund Tier 2 (Award Number: MOE-T2EP20122-0006).

\end{document}